\documentclass[sigconf,screen,authorversion,nonacm]{acmart}

\usepackage{booktabs}       
\usepackage{multirow}
\usepackage{comment}        
\usepackage{verbatim}       
\usepackage{algorithm, algorithmic}
\usepackage{makecell}
\usepackage{amsmath}
\usepackage{bm, bbm}
\usepackage{amsthm}
\usepackage{mathtools}
\newtheorem{theorem}{Theorem}[section]
\newtheorem{definition}{Definition}[section]

\newtheorem{assumption}{Assumption}[section]

\usepackage{enumitem}
\setlist{itemsep=0.2ex, topsep=1ex, partopsep=0ex, parsep=0.5ex, leftmargin=2.3ex}

\def\E{\mathbb{E}}

\def\R{\mathbb{R}}

\def\eps{\epsilon}
\def\indep{\perp \!\!\! \perp}
\def\notindep{\not\!\perp\!\!\!\perp}
\newcommand{\argmin}{\mathop{\mbox{argmin}}}

\usepackage{xcolor}
\definecolor{purple}{rgb}{0.3,0.0,.4}

\makeatletter
\def\Cline#1#2{\@Cline#1#2\@nil}
\def\@Cline#1-#2#3\@nil{%
  \omit
  \@multicnt#1%
  \advance\@multispan\m@ne
  \ifnum\@multicnt=\@ne\@firstofone{&\omit}\fi
  \@multicnt#2%
  \advance\@multicnt-#1%
  \advance\@multispan\@ne
  \leaders\hrule\@height#3\hfill
  \cr}
\makeatother

\AtBeginDocument{%
  \providecommand\BibTeX{{%
    \normalfont B\kern-0.5em{\scshape i\kern-0.25em b}\kern-0.8em\TeX}}}

\setcopyright{acmcopyright}
\copyrightyear{2023}
\acmYear{2023}
\acmDOI{XXXXXXX.XXXXXXX}

\acmConference[KDD '23]{the 29th ACM SIGKDD
Conference on Knowledge Discovery and Data Mining}{August 6 - 10, 2023}{Long Beach, CA}
\acmPrice{15.00}
\acmISBN{978-1-4503-XXXX-X/18/06}

\begin{document}

\title{The Missing Indicator Method: From Low to High Dimensions}

\author{Mike Van Ness}
\affiliation{%
  \institution{Stanford University}
  \city{Stanford}
  \state{CA}
  \country{USA}
}

\author{Tomas M. Bosschieter}
\affiliation{%
  \institution{Stanford University}
  \city{Stanford}
  \state{CA}
  \country{USA}
}
\author{Roberto Halpin-Gregorio}
\affiliation{
  \institution{Cornell University}
  \city{Ithaca}
  \state{NY}
  \country{USA}
}

\author{Madeleine Udell}
\affiliation{%
  \institution{Stanford University}
  \city{Stanford}
  \state{CA}
  \country{USA}
}

\renewcommand{\shortauthors}{Van Ness et al.}

\begin{abstract}
Missing data is common in applied data science, particularly for tabular data sets found in healthcare, social sciences, and natural sciences.
Most supervised learning methods only work on complete data, thus requiring preprocessing such as missing value imputation 
to work on incomplete data sets.
However, imputation alone does not encode useful information about the missing values themselves.
For data sets with informative missing patterns, 
the Missing Indicator Method (MIM), which adds indicator variables to indicate the missing pattern, can be used in conjunction with imputation to improve model performance.
While commonly used in data science, 
MIM is surprisingly understudied from an empirical and especially theoretical perspective.
In this paper, we show empirically and theoretically that MIM improves performance for informative missing values, and we prove that
MIM does not hurt linear models asymptotically for uninformative missing values.
Additionally, we find that for high-dimensional data sets with many uninformative indicators, MIM can induce model overfitting and thus test performance. 
To address this issue, we introduce Selective MIM (SMIM), a novel MIM extension that adds missing indicators only for features that have informative missing patterns.
%
We show empirically that SMIM performs at least as well as MIM in general, and improves MIM for high-dimensional data.
Lastly, to demonstrate the utility of MIM on real-world data science tasks, we demonstrate the effectiveness of MIM and SMIM on clinical tasks generated from the MIMIC-III database of electronic health records.
\end{abstract}




\maketitle

\section{Introduction}

Missing data is an unavoidable consequence of tabular data collection in many domains.  
Nonetheless, most statistical studies and machine learning algorithms not only assume complete data, but cannot run on data sets with missing entries. 
Two common preprocessing methods are used to address this issue. 
The first method is complete case analysis, in which 
the fully observed data samples are used, while the others are discarded.
However, discarding data reduces statistical power and can bias results, 
as partially-observed data samples may still contain important information.
Furthermore, sometimes \emph{all} data samples have missing values, in which case discarding incomplete samples is catastrophic. 

The second method is missing value imputation, in which 
missing values are replaced by estimates from the observed data.
While missing value imputation is well studied and widely used \cite{van2018flexible, lin2020missing}, it comes with two problems.
Firstly, most imputation methods are only (provably) effective when missing values are missing at random (MAR), which is often violated in real-world data, e.g. medical data \cite{perez2022benchmarking}. 
In particular, many real-world data sets exhibit \textit{informative missingness}, i.e. when the pattern of missing data encodes information about the response variable.
When data has informative missingness, the MAR assumption is not typically satisfied, making imputation methods less effective.

Secondly, imputation methods are typically evaluated by their reconstruction error.
This method of evaluation is reasonable when the task of interest itself is imputation, e.g. for recommender systems.
However, optimal imputation in terms of lowest reconstruction error is not necessarily required for optimizing downstream prediction accuracy.
In fact, recent theoretical work has shown that simple imputation schemes can still result in Bayes optimal predictions if the imputation function is paired with an appropriate prediction function \cite{josse2019consistency, le2021sa, bertsimas2021prediction}.
Thus, it is possible to obtain optimal predictions without accurate imputations, which is particularly useful when making accurate imputations is challenging, i.e. for MNAR data.

An alternative strategy, particularly when missing values are informative, is the Missing Indicator Method (MIM), which directly leverages the signal in the missingness itself rather than optimizing imputation accuracy. 
For each partially observed feature, MIM adds a corresponding indicator feature to indicate whether the feature is missing or not.
MIM can be used with any imputation method, but is most commonly used with mean imputation, which is equivalent to 0-imputation after centering features.
MIM is a common method in practice, e.g. it is implemented in scikit-learn \cite{scikit-learn}, yet MIM remains understudied from an empirical and, in particular, theoretical perspective.


In this paper, we show both theoretically and empirically that MIM is an effective strategy for supervised learning in a wide variety of data regimes.
While some previous work has studied MIM related to statistical inference \cite{knol2010unpredictable, groenwold2012missing}, 
our work focuses on MIM as part of a supervised learning pipeline, where prediction accuracy is the primary objective opposed to inference.
Additionally, to better handle high-dimensional data sets where extra uninformative indicator features can lead to overfitting, we introduce \textit{Selective MIM} (SMIM), a novel MIM extension to adaptively select which indicators are useful to add to the data set.



Our main contributions are as follows:
\begin{itemize}
    \item We provide a novel theoretical analysis of MIM for linear regression. In particular, we prove that 
    MIM successfully encodes the signal in missing values when the missingness is informative, thus increasing prediction performance.
    Further, when missingness is not informative, MIM does not degrade performance asymptotically.
    
    \item We introduce \emph{Selective MIM} (SMIM), a novel improvement to MIM that uses significance tests to select only the informative indicators. SMIM stabilizes MIM for high-dimensional data sets, where MIM can cause overfitting and thus increased variance in training.
    
    \item We conduct an extensive empirical study of MIM and SMIM on synthetic data and real-world data, for various different imputation and prediction models.
    We show that MIM plus mean imputation is a strong method while being much more efficient than more complicated imputation schemes.
    Additionally, we show that SMIM performs at least as well as MIM in general, while outperforming MIM on high-dimensional data sets.
    \item To demonstrate the utility of MIM and SMIM in real-world applications, we evaluate these methods on the MIMIC-III data set \cite{johnson2016mimic}, a clinical data set of electronic health records (EHRs).
    We show that (S)MIM improves model performance on a collection of clinical tasks, showing that real-world data often has informative missing values with signal that can be captured with MIM.
\end{itemize}

\section{Missing Values in Supervised Learning}\label{sec:setup}
In this section we introduce the problem statement and relevant work, and expand on how we build on existing work.

\subsection{Notation}

We use the following conventions.  Upper case letters $X$ are random variables.  
Bold letters $\bm{X}$ are vectors, which by default are always column vectors.  
Subscripts $X_j$ denote components of $\bm{X}$, and superscripts $\bm{X}^{(i)}$ denote sample vectors in a data set.  
We use $n$ for the number of samples/rows, and $p$ for the number of features/columns.  
We reserve $i$ as an index on samples and $j$ as an index on features.
$\bm{R}$ will denote indicator features corresponding to missing values in $\bm{X}$.  The sets $\text{obs}(\bm{R}) = \{j : R_j = 0\}$ and $\text{miss}(\bm{R}) = \{j : R_j = 1\}$ indicate observed and missing components of a vector.
$X_j \indep X_k$ means $X_j$ and $X_k$ are independent as random variables.


\subsection{Problem Statement}\label{sec:prob_state}
We consider the following supervised learning setup. 
Let $\bm{X} \in \R^p$ be a vector of $p$ features, and $Y \in \mathcal Y$ a response to be predicted from $\bm{X}$, where $\mathcal{Y} = \R$ for regression tasks and $\mathcal{Y} = \{0, 1, \ldots, k-1\}$ for k-class classification tasks. 
The vector $\bm{X}$ contains a complete set of values, in the sense that a value exists for each component of $\bm{X}$ even if that value is not observed. 
In reality, we observe $\bm{Z} \in \{ \R \cup \{*\}\}^p$ where $Z_j = X_j$ when $X_j$ is observed and $Z_j = \text{`}{}*{}\text{'}$ when $X_j$ is unobserved/missing.
$\bm{Z}$ yields the random binary vector $\bm{R} \in \{0, 1\}^p$ that indicates the missing and observed components of $\bm{Z}$. 
Specifically, $R_j = 0$ when $X_j$ is observed and $R_j = 1$ when $X_j$ is missing. 

Given a loss function $\mathcal L: \mathcal Y \times \mathcal Y \to \R$ and a distribution over pairs $(\bm{Z}, Y) \sim \mathcal{D}$, the goal is to find function $f: \{ \R \cup \{*\}\}^p \to \mathcal Y$ that minimizes the expected loss:
\begin{equation}
\label{eq:full_loss}
f^* = \argmin_{f} ~\E_{\mathcal D} \left[ \mathcal{L}(f(\bm{Z}), Y)\right].
\end{equation}
We will often drop the subscript $\mathcal{D}$ from expectations with the default being that the expectation is with respect to $\mathcal{D}$.
The function $f$ can be a pipeline that combines an imputation method with a prediction method, or can directly predict $Y$ from $\bm{Z}$ without imputation, e.g. fitting a different model per missing pattern. 

\subsection{Types of Missing Data}

Traditionally, partially observed data is categorized by the missing mechanism, i.e. the distribution of $\bm{R}$ conditional on the complete data $\bm{X}$.
The most typical categorization of missing mechanisms is
\cite{little2019statistical}:
\begin{itemize}
    \item \textbf{Missing Completely at Random (MCAR).} The missing pattern is random and independent of the data: $P(\bm{R}\mid\bm{X}) = P(\bm{R})$.
    \item \textbf{Missing at Random (MAR).} The missing pattern is independent of unobserved values $\bm{X}_{\text{miss}(\bm{R})}$ conditioned on observed values $\bm{X}_{\text{obs}(\bm{R})}$: $P(\bm{R} \mid \bm{X}) = P(\bm{R} \mid \bm{X}_{\text{obs}(\bm{R})})$.
    \item \textbf{Missing Not at Random (MNAR).} The missing pattern may depend on unobserved values.
\end{itemize}
The MCAR and MAR mechanisms are sometimes considered \emph{ignorable} in the sense that these mechanisms may be ignored for likelihood-based inference \cite{rubin1976inference}. 
Many imputation methods work well on MCAR or MAR data \cite{van2011mice, stekhoven2012missforest, honaker2011amelia, zhao2020missing}, while MNAR data is more challenging for imputation methods and requires modeling the missing mechanism itself \cite{sportisse2020estimation, mattei2019miwae, kyono2021miracle}. 

While missing mechanisms define the relationship between $\bm{R}$ and $\bm{X}$, informative missingness is defined by the relationship between $\bm{R}$ and $Y$:
\begin{definition}\label{def:inf_miss}
For $(\bm{Z}, Y) \sim \mathcal{D}$ with missing pattern $\bm{R}$, the missing pattern is \emph{informative} if $\bm{R} \notindep Y$.
\end{definition}
Assuming $\bm{X} \notindep Y$, this definition implies that MAR and MNAR data always have informative missingness. 
MCAR data can also have informative missingness if $\bm{R} \notindep Y$ despite $\bm{R} \indep \bm{X}$, although this is rare in practice.
In previous literature, informative missingness is most often associated with MNAR data, as in this setting the missingness is most informative \cite{schafer2002missing, li2018don}.
In general, we expect MIM to improve predictions when missingness is informative.
Yet, we can have MAR data and not see any benefit to MIM over imputation alone if $\bm{R} \indep Y \mid \bm{X}_{obs}$


\subsection{Related Work}
\label{sec:related_work}

Missing values have been an important topic of study in statistics for decades. 
Classical work focused on statistical inference problems \cite{rubin1976inference, dempster1977maximum, little2019statistical}, while more recent work has focused more on imputation, ranging from statistical methods \cite{honaker2011amelia, zhao2020matrix} to iterative approaches \cite{van2011mice, stekhoven2012missforest} and deep learning  \cite{yoon2018gain, mattei2019miwae, gondara2018mida}.

Even though MIM is commonly used in practice (e.g. it is implemented in sklearn \cite{scikit-learn}), 
its properties are surprisingly understudied in statistical and machine learning literature, particularly from the perspective of supervised learning.
MIM has been explored more in the medical literature due to the ubiquity of missing values in medical data sets, and particularly the tendency of these missing values to be informative.
A few medical papers advocate for MIM with mean imputation \cite{sharafoddini2019new, sperrin2020missing}, and some for MIM with other imputation methods \cite{sperrin2020multiple, qu2009propensity}. 
Other medical papers caution the use of MIM when statistical estimation and inference is the downstream task, as then using MIM can add bias to parameter estimates \cite{knol2010unpredictable, groenwold2012missing}.
We instead focus on the problem of optimal supervised learning.


Many previous methodological and empirical works have studied missing value preprocessing for supervised learning, yet most do not include MIM or a similar method to capture informative missingness.
Some empirical studies have evaluated combinations of imputation and prediction models \cite{garciarena2017extensive, woznica2020does}, but these most often exclude MIM.
Recent AutoML methods attempt to optimize the entire supervised learning pipeline \cite{yang2020automl, garciarena2017evolving, feurer2020auto}, but without considering MIM.
Deep learning can be used to jointly optimize imputation and prediction if both models are neural networks \cite{le2020neumiss, ipsen2021deal}, but these methods do not capture informative missingness without additionally using missing indicators.
Decision trees can uniquely handle missing values without imputation, e.g. using the Missing Incorporated as Attribute method \cite{twala2008good, josse2019consistency}, but such methods are only applicable to tree-based methods. 

Lastly, some recent papers have explored impute-then-predict pipelines from a theoretical perspective.
\cite{josse2019consistency, le2021sa, bertsimas2021prediction} show that accurate imputation is not needed to produce Bayes optimal predictions as long as a powerful enough predictor is used.
While this theory is not immediately applicable in practice, since constructing these predictors would often be impossible, it does give motivation for developing missing value preprocessing methods that do not focus on imputation accuracy, such as MIM with mean imputation.
Perhaps the most relevant previous work to this paper is \cite{le2020linear}, which gives a risk bounds for linear models with 0-imputation using both MIM and an expanded linear model. 
We also present theory for MIM with linear models in this paper, but under different assumptions and focusing on asymptotics rather than risk bounds. 
We also perform a much more diverse set of empirical experiments in this paper compared to \cite{le2020linear}.

\section{Missing Indicator Method Theory}\label{sec:theory}

In this section, we present a theoretical study of MIM for linear regression. 
Following the notation from Section \ref{sec:setup}, let $\bm{X} \in \R^p$ be $p$ features, $Y \in \R$ a continuous response, and $\bm{Z} \in \left\{ \R \cup \{*\}\right\}^p$ the observed features with potentially missing values. The full optimization problem in Eq.~\eqref{eq:full_loss} with squared error loss $\mathcal{L}(x, y) = (x - y)^2$ is solved by the conditional expectation $\E[Y \mid \bm{Z}]$. However, as explained in \cite{le2020linear, le2020neumiss}, this conditional expectation is combinatorial in nature, as it involves estimating a different expectation for each missing pattern $\bm{R}$:
\begin{align}
\label{eq:loss_cond_exp}
f^* = \E[Y \mid \bm{Z}] &= \E[Y \mid \bm{X}_{\text{obs}(\bm{R})}, \bm{R}]  \\
\label{eq:loss_comb}
&= \sum_{\bm{r} \in \{0, 1\}^d} \E[Y \mid \bm{X}_{\text{obs}(\bm{r})}, \bm{R} = \bm{r}] \mathbbm{1}_{\bm{R} = \bm{r}}.
\end{align}
In \cite{le2020linear, le2020neumiss}, the conditional expectation in Eq.~\eqref{eq:loss_cond_exp} is estimated after assuming that $\bm{X}$ comes from a Gaussian distribution. In this paper, we make no distributional assumptions,   and instead make the following assumptions:
\begin{assumption}
\label{as:center_and_impute}
$Y$ is centered as $Y \gets Y - \E[Y]$, while $\bm{Z}$ is centered over the observed entries $Z_j \gets Z_j - \E[Z_j \mid R_j = 0]$ and imputed with 0, resulting in the imputed data vector $\tilde{\bm{Z}}$ defined as
\begin{equation}
\label{eq:zero_impute}
 \tilde{Z}_j =
 \begin{cases}
X_j & R_j = 0, \\
0 & R_j = 1.
 \end{cases}
\end{equation}
\end{assumption}
\begin{assumption}
\label{as:blp}
$f$ is constrained to be a linear function, and thus $f^*$ is the best \textit{linear} prediction function.
\end{assumption}

We now examine $f^*$ with and without MIM. Specifically, we compare the best linear predictors (BLPs):
\begin{align}
\label{eq:no_mim_risk}
\bm{\beta}^* &= \argmin_{\bm{\beta}} \E [(Y - \tilde{\bm{Z}}^T \bm{\beta})^2], \\
\label{eq:mim_risk}
\bm{\beta}^*_{MIM}, ~\bm{\gamma}^*_{MIM} &= \argmin_{(\bm{\beta}, \bm{\gamma})} \E [(Y - \tilde{\bm{Z}}^T \bm{\beta} - \bm{R}^T \bm{\gamma})^2].
\end{align}
In practice, we would estimate these BLPs using a training set $\{(\bm{Z}^{(1)}, \bm{R}^{(1)}), \ldots (\bm{Z}^{(n)}, \bm{R}^{(n)})\}$, obtaining the standard ordinary least squares (OLS) estimates $\hat{\bm{\beta}}$, $\hat{\bm{\beta}}_{MIM}$, and 
$\hat{\bm{\gamma}}_{MIM}$.
Under mild conditions, most notably that $\E[\tilde{\bm{Z}} \tilde{\bm{Z}}^T]$ and $\E[(\tilde{\bm{Z}}^T, \bm{R}^T)^T (\tilde{\bm{Z}}^T, \bm{R}^T)]$ are invertible,
these OLS estimates 
will converge to the corresponding BLPs in Eqs.~\eqref{eq:no_mim_risk} and \eqref{eq:mim_risk}; see \cite{hansen2022econometrics} for further details.
This convergence validates the study of the best linear predictors, as they serve as the limits of the OLS estimates obtained in practice.





Theorem \ref{thm:1ft-case} starts by analyzing the simple case when $p = 1$, which has particularly nice properties.

\begin{theorem}\label{thm:1ft-case}
Grant Assumptions \ref{as:center_and_impute} and \ref{as:blp}.  $p = 1$, and let $\mathcal{M} = \{i \colon R^{(i)} = 1\}$ be the missing samples, then the OLS estimates are
\begin{equation}
\hat{\beta}_{MIM} = \hat{\beta}, \quad \hat{\gamma}_{MIM} = \frac{1}{|\mathcal{M}|} \sum_{i \in \mathcal{M}} Y^{(i)},
\end{equation}
and thus the BLPs are 
\begin{equation}
\beta^*_{MIM} = \beta^*, \quad \gamma^*_{MIM} = \E[Y \mid R = 1].
\end{equation}
\end{theorem}
A direct corollary of Theorem \ref{thm:1ft-case} is that if $R$ is uninformative, then
\begin{equation}
\gamma^*_{MIM} = \E[Y \mid R = 1] = \E[Y] = 0
\end{equation}
since $Y$ is centered. On the other hand, if $R$ is informative, then $\gamma^*_{MIM} = \E[Y \mid R = 1] \neq 0$, allowing the model to adjust its predictions when $X$ is missing.
Specifically, because $X$ is imputed with 0, the best linear prediction function is
\begin{equation}
f^*_{linear}(X) = 
\begin{cases}
X \beta^* & \textnormal{if } R = 0, \\
\E[Y \mid R = 1] & \textnormal{if } R = 1.
\end{cases}
\end{equation}
or correspondingly in finite sample
\begin{equation}
\hat{f}_{linear}(X) = 
\begin{cases}
X \hat{\beta} & \textnormal{if } R = 0, \\
\frac{1}{|\mathcal{M}|} \sum_{i \in \mathcal{M}} Y^{(i)} & \textnormal{if } R = 1.
\end{cases}
\end{equation}
We see that when $X$ is observed, the model ignores $R$ as a feature, and when $X$ is missing, the model predicts the average of $Y$ among the missing values. Note that when $p=1$, this result occurs both in finite sample and asymptotically.

We now consider the more general case of $p > 1$ features in Theorem~\ref{thm:ols_mcar}.

\begin{theorem} \label{thm:ols_mcar}
Grant Assumptions \ref{as:center_and_impute} and \ref{as:blp}.
\begin{enumerate}[label=(\alph*), leftmargin=3.5ex]
    \item If the missing mechanism is MCAR and $\bm{R}$ is uninformative, then 
    \begin{equation}
    \bm{\beta}^* = \bm{\beta}^*_{MIM}, \quad \bm{\gamma}^* = 0.
    \end{equation}
    
    \item If (i) the missing mechanism is self-masking, i.e.
    $P(\bm{R} \mid \bm{X}) = \prod_j P(R_j \mid X_j)$; (ii) $X_j \indep X_k$ for $j \neq k$; and (iii) $\bm{R}$ is centered, then
    \begin{gather}
    \bm{\beta}^* = \bm{\beta}^*_{MIM}, \\
    \end{gather}
    and for $j = 1, \ldots, p$
    \begin{equation}
    \label{eq:gamma_ols_mim}
    \gamma^*_{MIM_j} = \E[Y \mid X_j \text{ missing}] - \E[Y \mid X_j \text{ observed}].
    \end{equation}
    
    \item If $\{1, \ldots, p\}$ can be partitioned into $d$ blocks $B_1, \ldots, B_d$ with block-independence: $R_j, X_j \indep R_k, X_k$ when $j, k$ are in different blocks, then for $j \in B_{m}$
    \begin{align}
        \begin{split}
        \label{eq:eq_ols_block}
        \beta^*_{MIM_j} = \sum_{\ell \in B_{m}} &a \E[\tilde{Z}_j Y] ~+ \\
        &b (\E[Y \mid X_\ell \text{ missing}] - \E[Y \mid X_\ell \text{ observed}] )
        \end{split}
    \end{align}
    for $a$ and $b$ are functions of $(\tilde{\bm{Z}}, \bm{R})$, and the same for $\gamma^*_{MIM_j}$ but for different $a$ and $b$.
\end{enumerate}
\end{theorem}

The proofs of Theorems \ref{thm:1ft-case} and \ref{thm:ols_mcar} are given in Appendix \ref{sec:proofs}. Theorem \ref{thm:ols_mcar} (a) shows that when the missing mechanism is MCAR and $\bm{R}$ is uninformative, the best linear predictions are the same regardless of whether MIM is used or not.
Since MIM is not expected to improve model performance when missingness is not informative, it is promising that MIM does not decrease the accuracy of linear models in this setting, at least asymptotically. This is important because practitioners are often unaware of the missing mechanism of the data, which is difficult to test \cite{baraldi2010introduction}. It is therefore crucial to understand the effects of MIM across all missing mechanisms. This evidence that MIM does not hurt performance even in a worst-case scenario justifies its use when predictive performance is prioritized.


Parts (b) and (c) of Theorem~\ref{thm:ols_mcar} show that when missingness is informative, the best linear prediction function using MIM learns how the missing values impact the distribution of $Y$, thereby leveraging the signal in the missing values.
In part (b), under the fully-independent self-masking mechanism, the form of $\bm{\gamma}^*_{MIM}$ in Eq.~\eqref{eq:gamma_ols_mim} directly accounts for the informative missingness in each feature, i.e. $\E[Y \mid X_j 
 \textnormal{ missing}] - \E[Y \mid X_j \textnormal{ observed}]$.
Part (c) presents a more general case, where independence is only assumed blockwise. 
Nonetheless, the formula for the BLP in (c) still contains the same expression as in Eq.~\eqref{eq:gamma_ols_mim} from (b), which encodes the (potential) informative missingness of each feature. 

As discussed in Section~\ref{sec:related_work}, even though MIM is commonly used in practice, very little previous work has studied MIM theoretically, especially from the lens of supervised learning. 
A notable exception is \cite{le2020linear}, in particular Theorem 5.2, which presents an upper bound for the risk of the linear model with MIM, i.e. the expected loss in Eq.~\eqref{eq:mim_risk}. \cite{le2020linear} compares this risk bound to the risk bound for the ``expanded'' linear model, which fits a separate linear model for each missing pattern in Eq.~\eqref{eq:loss_comb}. 
Theorem~\ref{thm:ols_mcar} in our paper is complementary to these results, as our results specify the values for the best linear predictors under specific missing mechanisms. 
While the risk bounds in \cite{le2020linear} are useful for comparing the finite-sample risk of MIM compared to other models, our results present a more practical understanding of linear model coefficients with MIM, and how these coefficients can directly encode or ignore missingness depending on the missing mechanism.

\section{Selective MIM}\label{sec:Selective-MIM}

Observe that MIM adds a missing indicator for \emph{every} feature that is partially observed, regardless of the missing mechanism. 
Theorem \ref{thm:ols_mcar} (a) justifies this behavior asymptotically, since the uninformative indicators can be ignored as $n \to \infty$. 
However, in finite sample, and particularly for high-dimensional data, adding many uninformative indicators can cause overfitting and thus can hurt model performance. 

Keeping only the informative indicators from MIM would help stabilize MIM on high-dimensional data.
One strategy to do such selection would be to use a statistical test to discover the missing mechanism, and use this information to determine which indicators to keep. 
However, testing for the missing mechanism is challenging. Previous literature has suggested tests for MCAR, notably Little's MCAR test \cite{little1988test}, but these tests have been criticized \cite{baraldi2010introduction}. Testing for MAR or MNAR, on the other hand, is impossible, since it would require access to the unobserved data. From Definition~\ref{def:inf_miss}, however, we can instead test for informative missingness directly by testing the relationship between $\bm{R}$ and $Y$, and determine which indicators are worth including, instead of adding all.

We propose a novel strategy to selectively add indicators based on a statistical test, which we call \textit{Selective MIM} (SMIM). 
The overall procedure for SMIM is summarized in Algorithm \ref{alg:smim_alg}. 
Instead of testing the relationship between $R_j$ and $\bm{X}$, as a test for the missing mechanism would, SMIM tests the relationship between $R_j$ and $Y$, 
directly yielding signal from the informative missingness towards the response of each missing indicator.
%
Since we assume $Y$ is complete, there are no issues testing the relationship between $R_j$ and $Y$, as there would be when testing for the missing mechanism.
Further, if $X_j$ is correlated with $Y$, then testing the relationship between $R_j$ and $Y$ serves as a proxy for testing the relationship between $X_j$ and $R_j$, which is otherwise impossible if $X_j$ has missing values.

Specifically, for feature $j$, SMIM tests whether $R_j$ and $Y$ are independent. 
When $Y$ is continuous, we use a two-sample t-test comparing the $E[Y \mid R = 1]$ and $E[Y \mid R = 0]$.
When $Y$ is discrete, we use a chi-squared test of independence between $Y$ and $R_j$. 
To correct for multiple testing across the features, we use Benjamini-Hochberg p-value correction \cite{benjamini1995controlling} with false discovery rate (FDR) of $\alpha$. 
Since false negatives (not adding an indicator that is informative) are more harmful than false positives (adding an indicator that is not informative), we use a relatively high FDR of $\alpha = 0.1$ throughout the paper. See Algorithm~\ref{alg:smim_alg}.

\begin{algorithm}
\caption{Selective MIM}
\begin{algorithmic}[1]\label{alg:smim_alg}
    \STATE \textbf{Input:} Missing indicators $\bm{R}$, response $Y$, error rate $\alpha$.
    \STATE \textbf{Output: } Indicator indices to keep $\mathcal{I} \subseteq \{1, \ldots, p\}$.
    \STATE $\text{pvals} \gets [ ~]$
    \FOR{$j=1$ {\bfseries to} $p$}
        \IF{ $Y$ is continuous}
            \STATE $\text{pval} \gets \text{t-test}(Y \mid R = 0, ~Y \mid R = 1)$
        \ELSIF{$Y$ is categorical}
            \STATE $\text{pval} \gets \text{Chi2-test}(\text{Contingency-Table}(Y, R_j))$
        \ENDIF
        \STATE $\text{pvals}[j] \gets \text{pval}$
    \ENDFOR
    \STATE reject $ = \text{Benjamini–Hochberg}(\text{pvals}, \alpha)$
    \RETURN $\mathcal{I} = \{j : \text{reject}_j = \text{true}\}$.
\end{algorithmic}
\end{algorithm}


\section{Experiments}\label{sec:experiments}

\begin{figure}
    \centering
    \includegraphics[width=\linewidth]{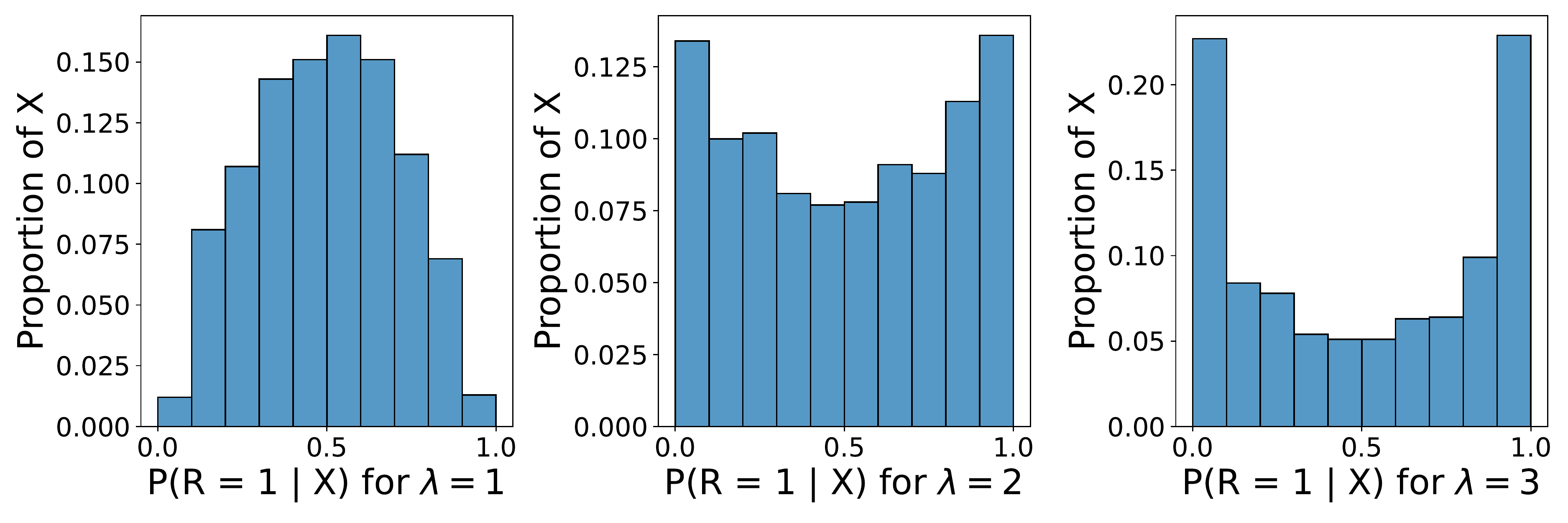}
    \vspace*{-5mm}
    \caption{Distributions of masking probabilities $P(R = 1 \mid X)$ generated from $X \sim \mathcal{N}(0, 1)$ using Eq.~\eqref{eq:self_mask} for different values of the informativeness parameter $\lambda$. Larger values of $\lambda$ result in `steeper' sigmoid functions in Eq.~\eqref{eq:self_mask}, hence more values lie in the masked-with-high-probability regime.}
    \label{fig:gamma_hists}
\end{figure}

\begin{figure}[t]
    \centering
    \includegraphics[width=\linewidth]{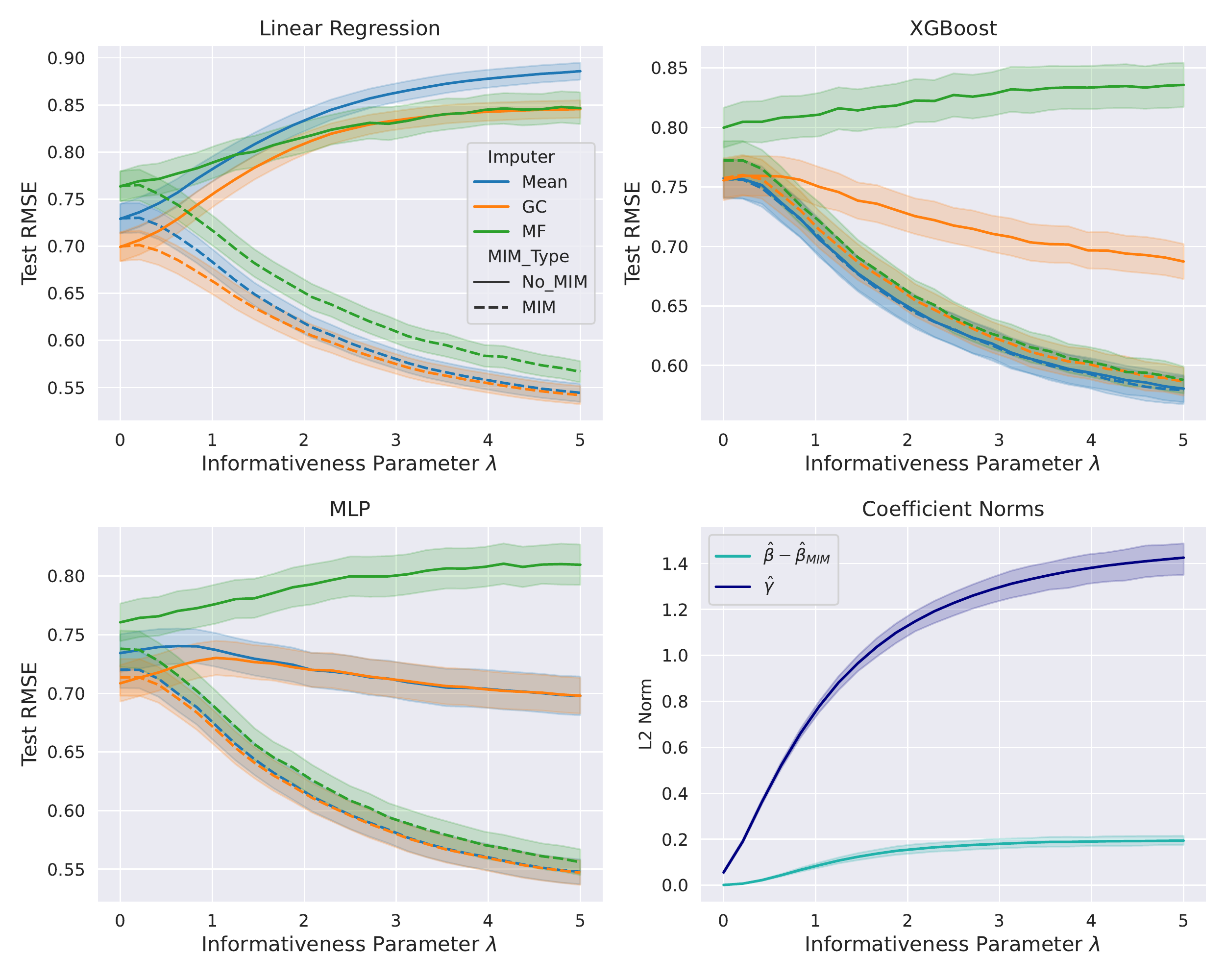}
    \vspace*{-5mm}
    \caption{MIM performance on synthetic data with $(n, p) = (10000, 10)$ with different imputation methods, as a function of the informativeness parameter $\lambda$ (see Eq.~\eqref{eq:self_mask}). 
    The top left, top right, and bottom left plots demonstrate that MIM (dashed lines) reduces test RSME in almost all scenarios when missingness is informative.
    The bottom right plot shows the linear regression coefficient norms, where $||\hat{\bm{\gamma}}||$ increases as missing value become more informative, as shown in Theorem \ref{thm:ols_mcar}.
    }
    \label{fig:sim_reg_low_dim}
\end{figure}

\begin{figure}[t]
    \centering
    \includegraphics[width=\linewidth]{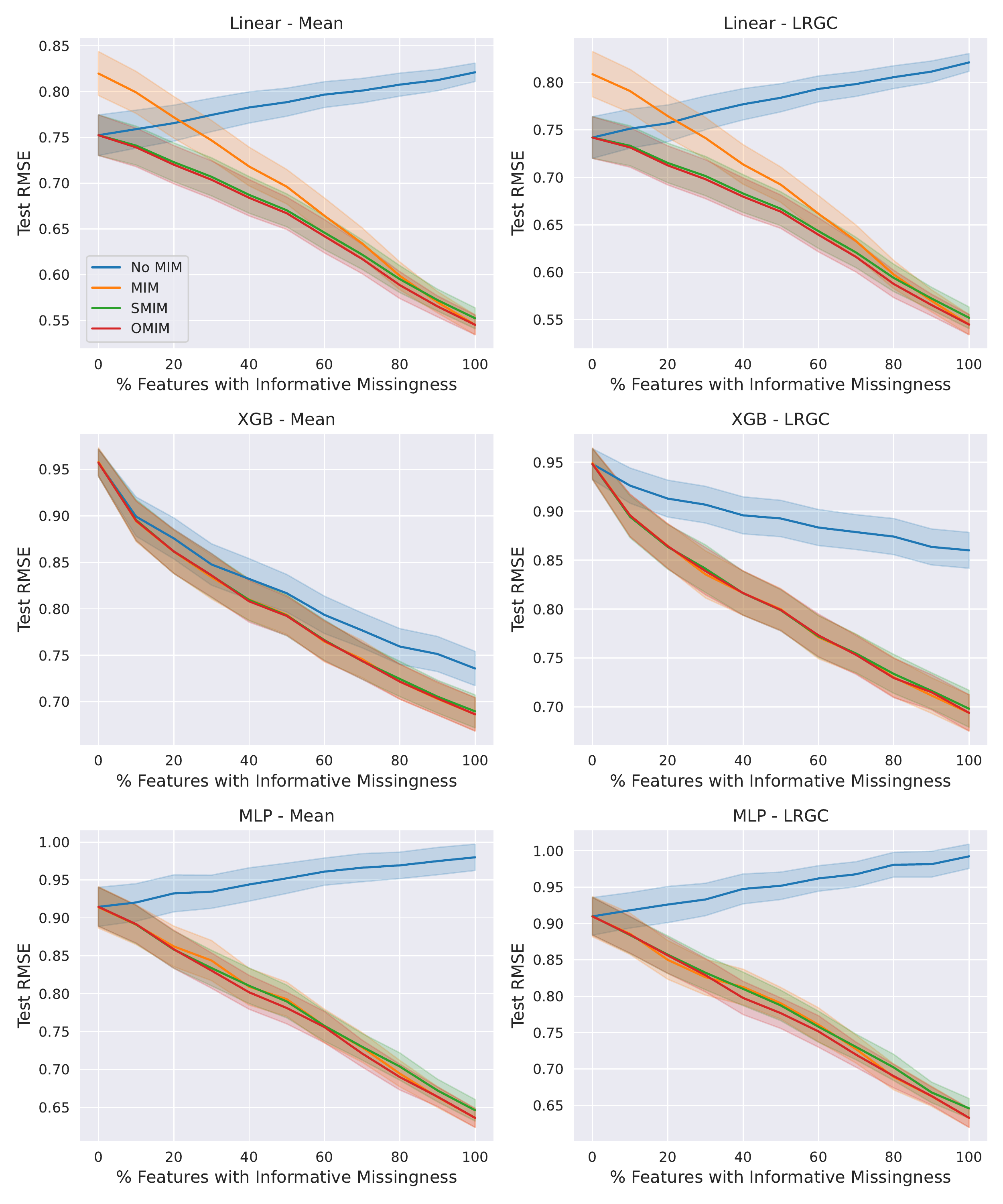}
    \vspace*{-5mm}
    \caption{
    Comparison of MIM, SMIM, Oracle MIM (OMIM), and No MIM
    on high-dimensional synthetic data with $(n, p) = (10000, 1000)$, as a function of the percent of features with informative missingness.
    }
    \label{fig:sim_smim}
\end{figure}

\begin{figure*}
    \centering
    \includegraphics[width=\linewidth]{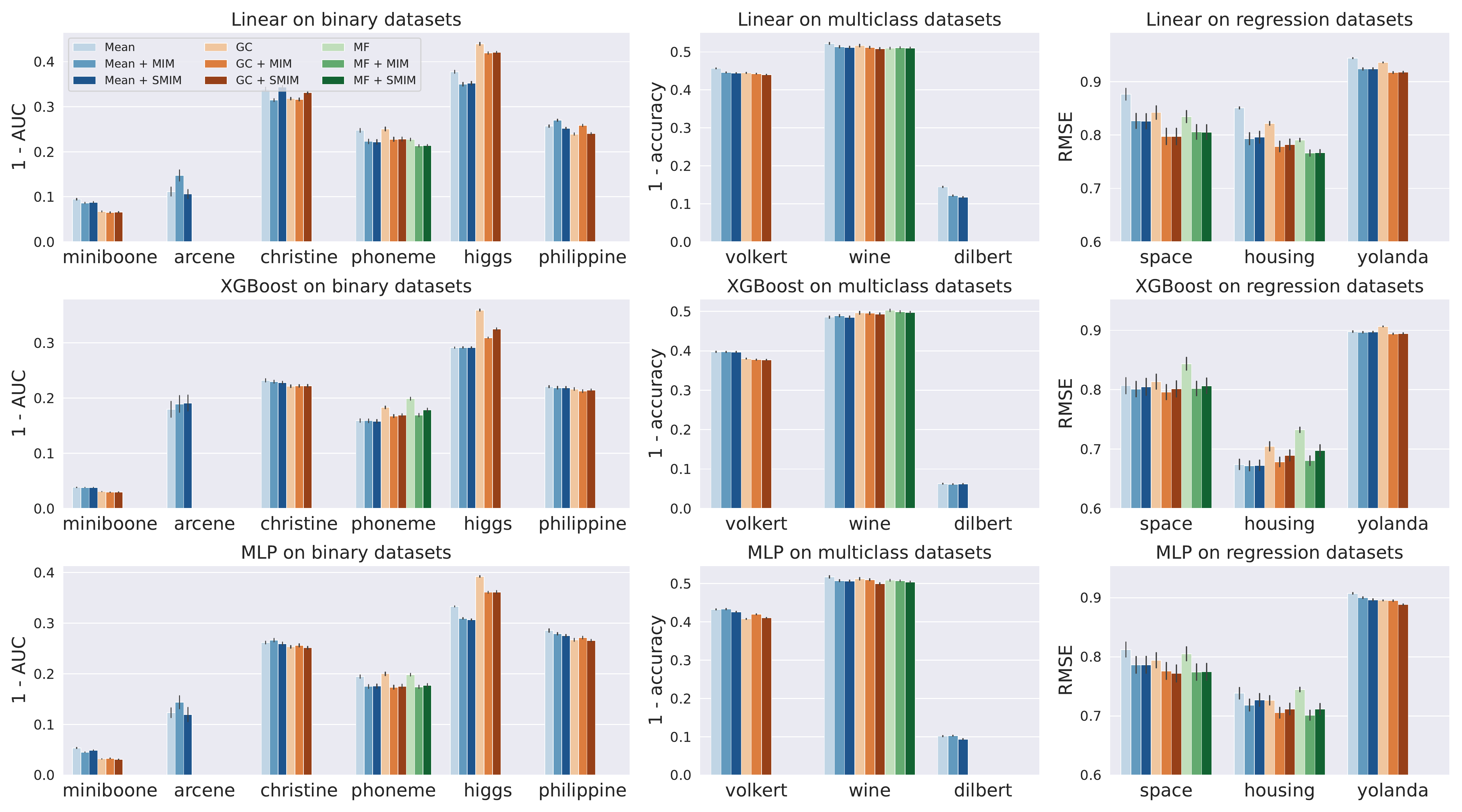}
    \caption{Test performance (lower is better) of MIM and SMIM with various imputation methods on OpenML data sets. Missing values are generated according to Eq.~\eqref{eq:self_mask}, where for each feature $\lambda_j = 2$ with probability $p_{\text{inf}} = 0.5$ and $\lambda_j = 0$ otherwise. Performance metric is RMSE for regression problems, 1-AUC for binary classification problems, 1-accuracy for multi-class problems. 
    Each bar represents the mean across 20 trials, and a
    a missing bar indicates a $> 3$ hour run time.
    }
    \label{fig:openml_main}
\end{figure*}

\begin{figure*}
    \centering
    \includegraphics[width=\linewidth]{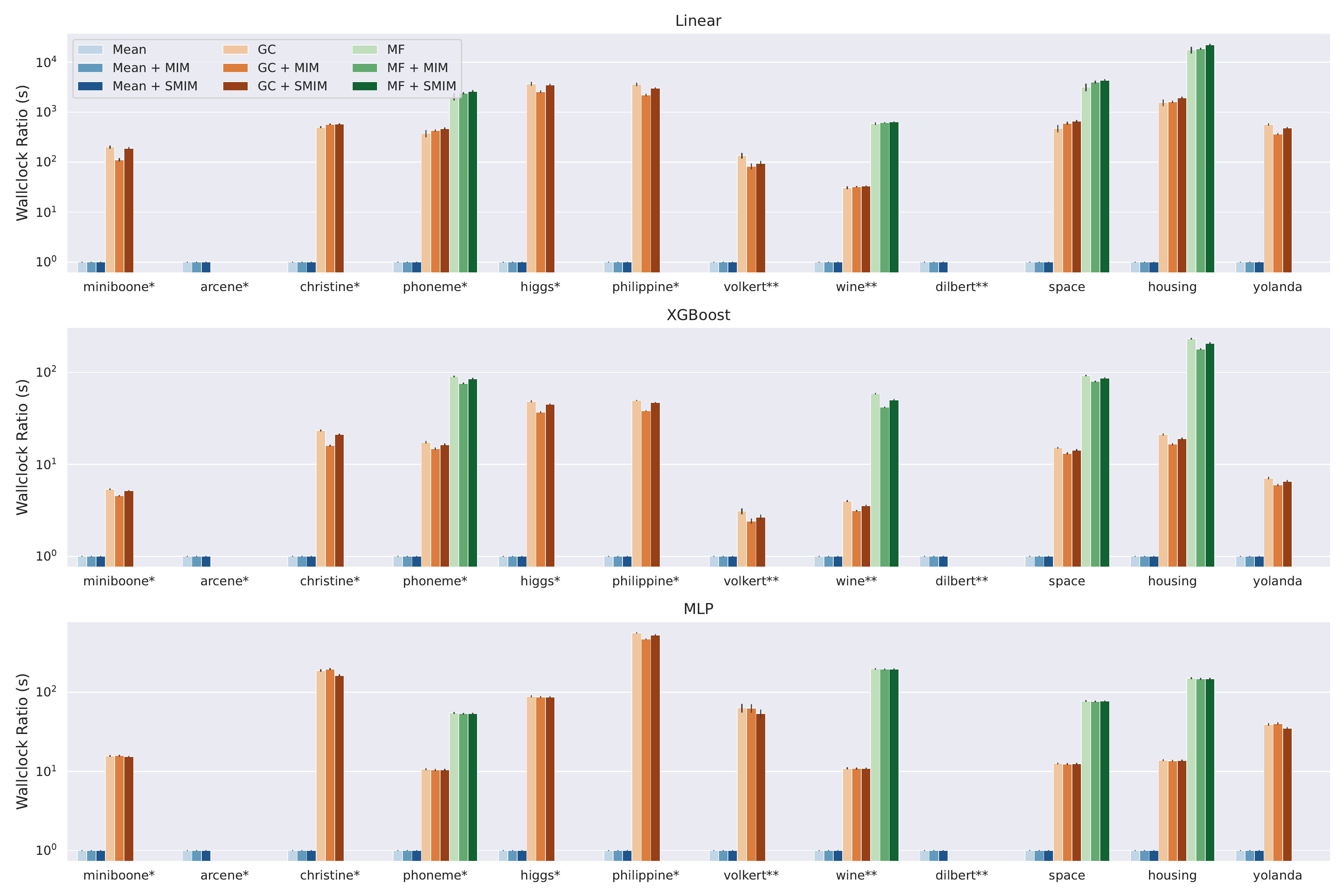}
    \caption{Wallclock times (in seconds) for results in Figure \ref{fig:openml_main}. 
    Each time is the ratio between the indicated method and the corresponding mean imputation method. 
    A missing bar indicates a $> 3$ hour run time, as in Figure \ref{fig:openml_main}.
    }
    \label{fig:openml_times}
\end{figure*}


We now empirically evaluate MIM and SMIM through experiments on synthetic data as well as real-world OpenML data sets with synthetic missing values.
We show that MIM boosts performance when missing values are informative, and does not negatively affect performance on most data sets with uninformative missing values.
Further, we show that SMIM can successfully discover the missing indicators that are informative, and is more effective than MIM on high-dimensional data sets. 

\subsection{Setup}\label{sec:exp_setup}

For all experiments in this section, we start with complete data and mask values, controlling for how much the missing pattern is informative. 
To generate the missing masks, we use a self-masking mechanism:
\begin{equation}
\label{eq:self_mask}
\begin{aligned}
P(R_j = 0 \mid \bm{X}) &= P(R_j = 0 \mid X_j) \\
&= \frac{1}{1 + \exp(-\lambda_j X_j)}.
\end{aligned}
\end{equation}
We call $\lambda_j$ the \textit{informativeness parameter} as it directly controls how informative the missing values in $X_j$ are by determining the steepness of the sigmoid masking probability function. The higher the value of $\lambda_j$, the more informative the missing values are. 
If $\lambda_j = 0$ for all $j$, then the missing values are MCAR and the missingness is uninformative. The impact of $\lambda_j$ on $R_j$ is showcased in more detail in Figure~\ref{fig:gamma_hists}.

We consider the following preprocessing methods to treat missing values:
\begin{itemize}
    \item \textbf{Mean: } Mean imputation over the observed entries. This is equivalent to 0-imputation, since we always standardize features.
    \item \textbf{MF: } Imputation via missForest \cite{stekhoven2012missforest}, an iterative imputer based on random forests \cite{breiman2001random}.
    \item \textbf{GC: } Imputation via gcimpute \cite{zhao2020missing, zhao2022gcimpute}, an EM-style imputer which uses the Gaussian Copula.
    \item \textbf{LRGC: } Imputation via gcimpute with the low rank Gaussian Copula model \cite{zhao2020matrix}, useful for high-dimensional data sets.
    \item \textbf{Imputer + MIM: } Imputation via the given imputer, along with MIM.
    \item \textbf{Imputer + SMIM: } Imputation via the given imputer, along with Selective MIM as in Algorithm \ref{alg:smim_alg}.
\end{itemize}
We occasionally use the term ``No MIM'' to refer to imputation without MIM or SMIM.
We choose these imputation methods to include either an iterative approach (MF) and an EM-based approach (GC), both popular classes of imputation methods.
For supervised learning models, we use linear/logistic regression, XGBoost \cite{chen2016xgboost}, and a multi-layer perceptron (MLP), giving us one model from the 3 most popular classes of supervised learning models (linear models, boosted decision trees, and neural networks). 
We standardize features over the observed entries in all experiments. 
We use a 75\%-25\% train-test data split and run each experiment for 20 trials, each with a different model seed, train-test split, and missing value mask.
To isolate the impact of the missing value handling, we analyze how different missing value preprocessing methods impact each supervised learning method separately, rather than comparing the supervised learning methods to each other. 
For performance metrics, we use RMSE for regression tasks, 1 - AUC for binary classification tasks, and 1 - accuracy for multiclass classification tasks (so in all cases \textit{lower is better}).
We release the code necessary to reproduce our results
\footnote{\url{https://github.com/mvanness354/missing_indicator_method}.}.

\subsection{MIM on Tree-Based Models}
\label{sec:mim_on_trees}
Before discussing our empirical results, we preface our findings with intuition for why XGBoost will likely behave differently with MIM than linear models and MLPs.
Tree-based methods treat all features as discrete, and so they discretize continuous features using binning \cite{breiman2017classification}. 
When using a constant imputation method like mean imputation, all imputed values always fall into the same bin and will always be split together by each tree.
We thus expect MIM to provide little additional information to the tree-based methods in this setting, since splitting on the indicator feature can be alternatively achieved by splitting twice to isolate the bin with the constant imputation value.
Meanwhile, if a non-constant imputation method is used, then MIM does add new splits for the tree to search over and thus has high potential to be valuable.
On the other hand, linear and neural network models do not treat continuous features as discrete, and thus are expected to benefit from missing indicators with all imputation methods.
A more detailed discussion of missing values in decision trees can be found in Appendix \ref{sec:mim_trees_appendix}.

\subsection{Synthetic Data} \label{sec:sim_data}

\paragraph{Low Dimensional Data}
We first consider the effects of MIM on synthetic data as a function of the informativeness parameter $\lambda$. 
Using $n = 10000$ and $p = 10$, we generate $\bm{X} \sim \mathcal{N}(0, \Sigma)$ with $\Sigma = \rho  \bm{1} \bm{1}^T + (1 - \rho)\mathcal{I}$ using $\rho = 0.3$ and $Y = \bm{X}^T \bm{\beta} + \eps$ for $\bm{\beta} \sim \mathcal{N}(0, \mathcal{I})$ and $\eps \sim \mathcal{N}(0, \sigma^2)$ with $\sigma^2$ chosen to enforce a signal-to-noise ratio of 10. 
We mask each feature according to Eq.~\eqref{eq:self_mask} with $\lambda_j = \lambda$ for all $j = 1, \ldots, p$.
The results for $\lambda \in [0, 5]$ are shown in Figure \ref{fig:sim_reg_low_dim}.
The top left, top right, and bottom left plots show that MIM continually reduces RSME as $\lambda$ increases across all imputation methods, which confirms that MIM is an effective preprocessor for informative missing values.
The lone exception is XGBoost, which benefits from MIM using GC and MF imputation but not when using mean imputatation.
This might be because mean imputation already allows trees to capture the informative signal in the missing values without MIM, as explained in Section~\ref{sec:mim_on_trees}.

The bottom right plot shows the linear regression coefficient norms $||\hat{\bm{\beta}} - \hat{\bm{\beta}}_{MIM}||_2$ and $||\hat{\bm{\gamma}}||_2$ as a function of $\lambda$, using the definition of these coefficients from Section \ref{sec:theory}. When $\lambda = 0$, both norms are close to 0, which we expect from Theorem \ref{thm:ols_mcar} (a) as it tells us that $\hat{\bm{\beta}} \approx \hat{\bm{\beta}}_{MIM}$ and $\hat{\bm{\gamma}} \approx 0$ when missingness is uninformative.
As $\lambda$ increases, i.e. the missing values become more informative, $||\hat{\bm{\gamma}}||_2$ increases while $||\hat{\bm{\beta}} - \hat{\bm{\beta}}_{MIM}||_2$ remains small. This behavior parallels Theorem \ref{thm:ols_mcar} (b), which shows that under a self-masking mechanism like Eq. (\ref{eq:self_mask}), $||\hat{\bm{\beta}} - \hat{\bm{\beta}}_{MIM}||_2$ should remain small but $||\hat{\bm{\gamma}}||_2$ should increase to capture the informative signal in each variable, i.e. $\E[Y \mid R_j = 1] - \E[Y \mid R_j = 0]$. 
It is noteworthy that this behavior still holds even without independence among the features in $\bm{X}$, suggesting that this behavior might hold, at least approximately, in more general linear regression settings.

\paragraph{High Dimensional Data}
We now consider high-dimensional synthetic data with $n = 10000$ and $p = 1000$. 
To generate realistic high-dimensional data, we generate $\bm{X} \sim \mathcal{N}(0, \Sigma)$ where $\Sigma$ is block diagonal with $b$ blocks and $d$ features per block, with block elements $\rho  \bm{1} \bm{1}^T + (1 - \rho)\mathcal{I}$ using $\rho = 0.5$.
We compute $\bm{X}^* \in \R^b$ as the block-wise mean of $\bm{X}$ (averaging over features in each block), and generate $Y$ as in the low dimensional case: $Y = \bm{X}^{*T} \bm{\beta} + \eps$ with $\eps \sim \mathcal{N}(0, \sigma^2)$.
This simulates real-world high-dimensional data where several features may be correlated with each other (but predominantly independent of other features), and thus are more likely to be missing together under informative missingness.

To generate the mask, we select only a random subset of blocks to have informative missingness: for each block, with probability $p_{\text{inf}}$ we set $\lambda=2$ for all features in the block as a whole, and with probability $1 - p_{\text{inf}}$ we set $\lambda=0$ for all features in said block.
In this setting, SMIM should be able to detect which features have informative missing values and only add the correponding indicators.
Along with MIM and SMIM, we also run experiments with Oracle MIM (OMIM), in which only features masked with $\lambda_j = 2$ are added, to represent an unrealistic but optimal preprocessor.

The results for $p_{\text{inf}} \in [0, 1]$ are shown in Figure \ref{fig:sim_smim}. 
We use LRGC instead of GC since the data is high-dimensional, and we do not report MF because computations do not terminate in less than 3 hours.
Using linear models, SMIM achieves comparable RMSE to OMIM and significantly better RMSE than No MIM across all values of $p_{\text{inf}}$. Further, SMIM outperforms MIM for smaller values of $p_{\text{inf}}$, when MIM adds many uninformative indicators. For MLP models, MIM and SMIM achieve comparable error to OMIM across all $p_{\text{inf}}$, demonstrating that MLPs can more readily ignore uninformative feature than linear models. Lastly, XGBoost with mean imputation is comparable for preprocessors, but suffers from No MIM with LRGC, further supporting the discussion on tree-based models in Section \ref{sec:mim_on_trees}.

\subsection{Masked Real-World Data}\label{sec:real_data}

We now run experiments on fully-observed real-world data sets obtained from the OpenML data set repository \cite{vanschoren2014openml}.
We mask entries according to Eq.~\eqref{eq:self_mask}, using $\lambda_j = 2$ with probability 0.5 and $\lambda_j = 0$ with probability 0.5 for all features $j$.
We select 12 data sets that cover a diverse spectrum of values for $n$ (number of samples), $p$ (number of features), and outcome type (binary, multiclass, regression); see Appendix \ref{sec:openml_dataset_stats} for further OpenML data set descriptions.
We focus on data sets with continuous features, although we discuss how MIM can be used with categorical features in Appendix \ref{sec:cat_features}.
Lastly, we use LRGC for high-dimensional data sets (arcene, christine, philippine, volkert, dilbert, yolanda) and use GC for all other data sets.

Figure \ref{fig:openml_main} shows the performance of each missing value preprocessing method paired with the 3 supervised learning models. 
For Linear and MLP models, MIM and SMIM improve performance across imputation methods for almost all data sets. This affirms the importance of MIM when missing values are informative.
XGBoost generally does as well with mean imputation as with other imputation methods, and is less impacted by MIM, supporting the discussion from Section \ref{sec:mim_on_trees}.
On high-dimensional data sets, SMIM outperforms MIM in most cases. This further demonstrates the value of discarding uninformative features in high-dimensional data, which SMIM can do effectively.

To better understand the time efficiency of the methods employed, we plot the relative wallclock times of each result from Figure \ref{fig:openml_main} in Figure \ref{fig:openml_times}. 
It is clear from Figure \ref{fig:openml_times} that the choice of imputation method dominates the computation time, instead of whether or not MIM or SMIM is used.
Specifically, mean imputation is orders of magnitude faster than GC and MF imputation. Also, MIM and SMIM appear to add relatively very little time to the total computation time, even though MIM doubles the number of features, suggesting that the method of imputation is much more important than the inclusion of MIM.
Further, notice that in Figure \ref{fig:openml_main}, (S)MIM with mean imputation often performs very comparably to (S)MIM with GC and MF.
Therefore, (S)MIM with mean imputation is an effective yet remarkably efficient alternative to using expensive imputation models.

\section{MIM on the MIMIC Benchmark}
\label{sec:mimic}

\begin{figure}[!h]
    \centering
    \includegraphics[width=\linewidth]{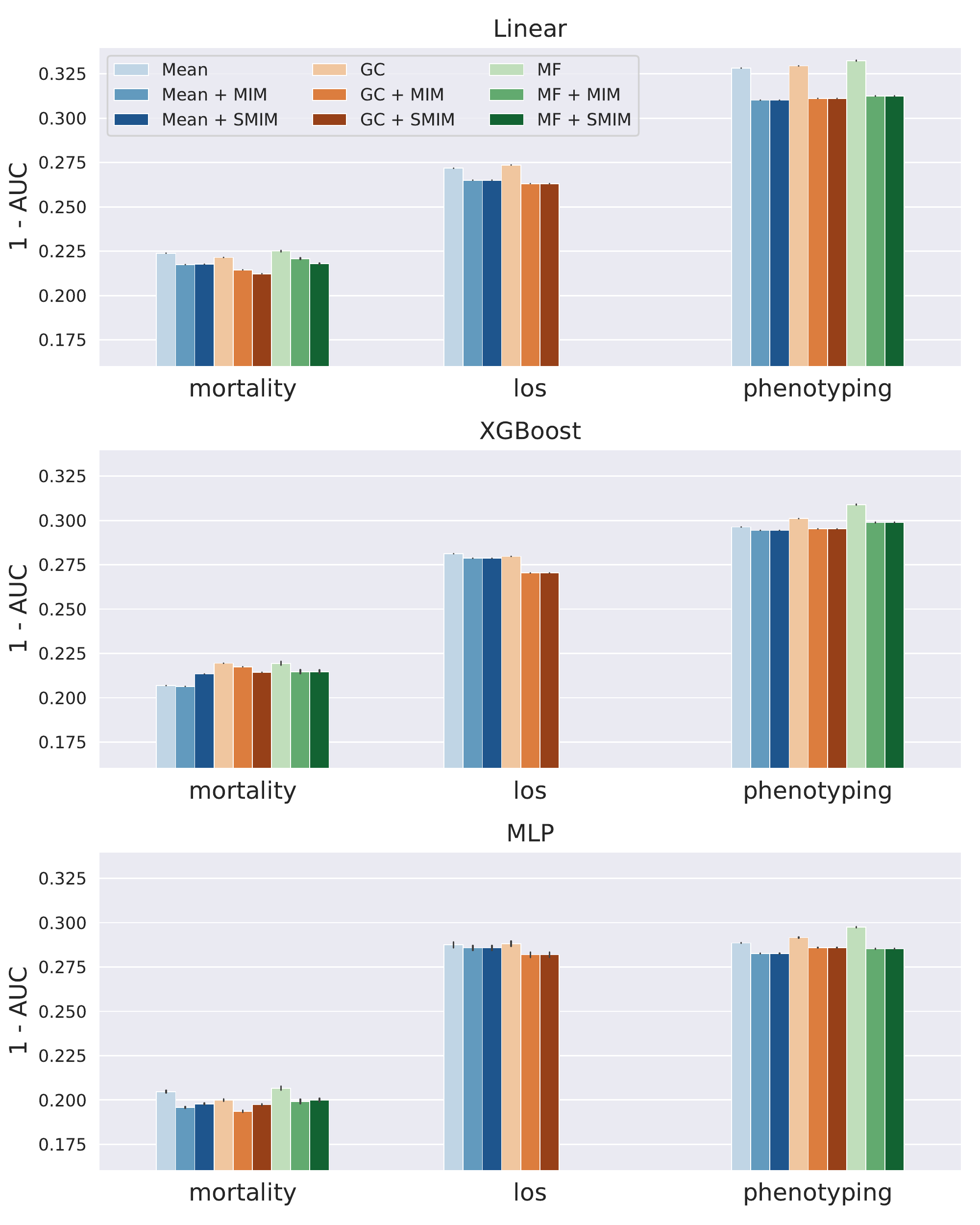}
    \caption{MIM and SMIM performance (lower is better) for clinical tasks on the MIMIC-III data set. Each bar represents the mean across 20 trials, each with a different random seed. A missing bar indicates a $>3$ hour run time.
    MIM and SMIM improve performance for logistic regression and MLP models on all 3 tasks, following the behavior exhibited in Figure \ref{fig:openml_main}.
    }
    \label{fig:mimic_plot}
\end{figure}

In Section \ref{sec:experiments}, we demonstrate that MIM and SMIM are effective tools for capturing informative missingness in synthetic and real-world data sets with synthetic missing values.
In this section, we demonstrate the effectiveness of MIM and SMIM on healthcare data which already has missing values.
Specifically, we use MIMIC-III \cite{johnson2016mimic}, an open source clinical database of electronic health records (EHRs) from the Beth Israel Deaconess Medical Center in Boston, Massachusetts.
To preprocess the MIMIC data, we utilize the mimic3-benchmark \cite{Harutyunyan2019}, which generates tabular data for various clinical tasks. 
We consider the following tasks for our experiments:
\begin{itemize}
\item \textbf{In Hospital Mortality Prediction}: binary mortality prediction using the first 48 hours of an ICU stay.
\item \textbf{Length-of-Stay Prediction (LOS)}: prediction of the remaining length-of-stay in the hospital. We formulate this as a binary prediction task to predict whether or not a patient will leave the hospital in the next 7 days.
\item \textbf{Phenotyping}: prediction of the acute care conditions present during a given hospital stay, formulated as a multi-label binary classification problem for 25 phenotypes. 
\end{itemize}
For each task, the target metric is 1 - AUC, following Section \ref{sec:experiments} and using a metric such that lower is better (for consistency across response types).
For phenotyping, we average the AUC for each of the 25 labels to create the macro AUC score, then set the metric as 1 - macro AUC. 
The train and test splits are established by the benchmark and kept constant across all trials.
The data set for each task contains 17 clinical variables that describe patient vital signs and other typical hospital measurements.
For additional details about our these variables and well as our experimental setup for MIMIC experiments, see Appendix \ref{sec:mimic_details}.

Figure \ref{fig:mimic_plot} shows the results of MIM and SMIM with the prediction pipelines used in Section \ref{sec:experiments}.
As on synthetic data and OpenML data, MIM and SMIM both consistently improve predictive performance with linear and MLP models, as well as with XGBoost when using GC or MF for imputations.
This result is particularly significant because it demonstrates that real-world data sets often have informative missing values, and MIM can help predictive models learn the signal in these missing values.
Additionally, there are no scenarios MIM negatively impacts predictive performance, confirming that MIM should be a standard preprocessing tool for supervised learning on low-dimensional real-world data sets, should predictive performance be prioritized.
In Figure \ref{fig:openml_realmiss} in Appendix \ref{sec:openml_realmiss}, we provide additional evidence that real-world data often contains informative missing values through additional experiments on OpenML data sets.

\section{Discussion}
\label{sec:discussion}

From our experiments on synthetic data, real-world data with synthetic missing values, and real-world EHR data, we obtain the following main takeaways:
\begin{itemize}
    \item When missing values are informative, MIM increases predictive performance of linear models and neural networks. 
    This is supported by Figure \ref{fig:sim_reg_low_dim}, where increasing the informativeness in synthetic data gradually increases the effectiveness of MIM; in OpenML data with informative missing values in Figure \ref{fig:openml_main}; and in EHR data in Figure \ref{fig:mimic_plot}. 

    \item Tree-based models in general benefit less from MIM than linear models and neural networks, particularly when using mean imputation. 
    We provide a possible explanation for this behavior in Section \ref{sec:mim_on_trees},
    and see this behavior manifest across all of our experiments.

    \item The only scenario where MIM might harm predictive performance is on high-dimensional data sets, e.g. the synthetic data in Figure \ref{fig:sim_smim} and on select high-dimensional OpenML data sets in Figure \ref{fig:openml_main}. 
    In these cases, Selective MIM (SMIM) is a stable extension of MIM that adds all the indicator features that have informative missing values.

    \item MIM and SMIM can increase performance not only with mean imputation, but also with other imputation methods, as shown in our experiments.
    Nonetheless, Figure \ref{fig:openml_times} shows that MIM with mean imputation is many orders of magnitude faster than using other imputation methods, yet usually results in about the same predictive performance.
    Therefore, MIM with mean imputation is a very effective and yet efficient way to treat missing values, especially under informative missingness when expensive imputation methods often bring little benefit.
    
\end{itemize}

\section{Conclusion}
\label{sec:conclusion}

When dealing with missing data, imputation with state-of-the-art imputers is often expensive. 
We show that using MIM in conjunction with mean imputation is an effective and efficient alternative, which we demonstrate via novel theory and comprehensive experimentation on synthetic data as well as real data with both synthetic and actual missing values.
We additionally introduce Selective MIM (SMIM), a MIM-based preprocessor that discards uninformative missing indicators and is thus more stable than MIM on high-dimensional data sets.
We show experimentally that adding MIM or SMIM helps achieve substantially better accuracy on data with informative missingness overall, and SMIM outperforms MIM on high-dimensional data. 
Future work might include researching theoretical guarantees on the use of imputation methods along with missing indicators, building on our theory.


\newpage
\bibliographystyle{ACM-Reference-Format}
\bibliography{bib}

\appendix

\section{Proofs}
\label{sec:proofs}

\subsection{Proof of Theorem~\ref{thm:1ft-case}}

Let $\bm{D} = [\tilde{Z}, R]^T$, then
\begin{equation}
\bm{D} \bm{D}^T = 
\begin{bmatrix}
\tilde{Z} \\
R
\end{bmatrix}
\begin{bmatrix}
\tilde{Z} & R
\end{bmatrix} = 
\begin{bmatrix}
\tilde{Z}^2 & 0 \\
0 & R
\end{bmatrix}.
\end{equation}
as $R^2 = R$ and $\tilde{Z}R = 0$ by zero imputation (Assumption \ref{as:center_and_impute}).
Thus the finite sample OLS estimates $\hat{\beta}$ and $\hat{\gamma}$ are
\begin{align}
\begin{bmatrix}
\hat{\beta}_{MIM} \\
\hat{\gamma}_{MIM}
\end{bmatrix} &= \E_n \left[ \bm{D}\bm{D}^T \right]^{-1} \E_n\left[ \bm{D} Y\right] \\
&=
\begin{bmatrix}
\E_n\left[ \tilde{Z}^2\right]^{-1} \E_n[\tilde{Z} Y] \\
\E_n\left[ R \right]^{-1} \E_n[R Y]
\end{bmatrix} \\
&= 
\label{eq:proof_1ft_finaleq}
\begin{bmatrix}
\hat{\beta} \\
\E_n[Y \mid R = 1]
\end{bmatrix}
\end{align}
The result for $\gamma$ in (\ref{eq:proof_1ft_finaleq}) holds because 
\begin{align}
\E_n\left[ R \right]^{-1} \E_n[R Y] &= \left( \frac{|\mathcal{M}|}{n}\right)^{-1} \frac{1}{n} \sum_{i \in \mathcal{M}} Y^{(i)} \\
&= \frac{1}{|\mathcal{M}|} \sum_{i \in \mathcal{M}} Y^{(i)} \\
&= \E_n [Y \mid R = 1]
\end{align}
\qed

\subsection{Proof of Theorem~\ref{thm:ols_mcar}}\label{sec:proof_thm_3.1}

\paragraph{Part (a)}
Let $\bm{D} = [\tilde{\bm{Z}}^T, \bm{R}^T]^T$
and $p_j = P(R = 1)$.
We would like to show that $\E\left[\bm{D} \bm{D}^T\right]$ is a $2 \times 2$ block diagonal matrix. When $j \neq k$, $\E[\tilde{Z}_j R_k] = 0$
since $X_j$ and $R_k$ are independent under MCAR.  If $j = k$, we cannot assume that $\tilde{Z}_j$ and $R_j$ are independent since they are directly dependent by construction.  Nonetheless, by the law of total expectations we have
\begin{align}
\E[\tilde{Z}_j R_j] &= \E[\tilde{Z}_j R_j \mid R_j = 1]p_j + \E[\tilde{Z}_j R_j \mid R_j = 0] (1 - p_j) \\
\label{eq:proof_total_exp}
&= \E[\tilde{Z}_j \mid R_j = 1]p_j \\
&= 0
\end{align}
since $\tilde{\bm{Z}}$ is imputed with 0.
We have now shown that $\E[\tilde{Z}_j R_k] = 0$ for all $j, k$, showing as desired that $\E\left[\bm{D} \bm{D}^T\right]$ is block diagonal. 
Further, since $R$ is uninformative, we have for all $j$
\begin{equation}
\E[R_j Y] = \E[R_j] \E[Y] = 0
\end{equation}
using the centering of $Y$. Thus we have
\begin{align}
\begin{bmatrix}
\bm{\beta}^*_{MIM} \\
\bm{\gamma}^*_{MIM}
\end{bmatrix} &= \left( \E[\bm{D} \bm{D}^T]\right)^{-1} \E[\bm{D} Y] \\
&=
\begin{bmatrix}
\E[\tilde{\bm{Z}}\tilde{\bm{Z}}^T]^{-1} & 0 \\
0 & \E[\bm{R} \bm{R}^T]^{-1}
\end{bmatrix}
\begin{bmatrix}
\E[\bm{\tilde{Z}} Y] \\
0
\end{bmatrix} \\
&= 
\begin{bmatrix}
\bm{\beta}^* \\
0
\end{bmatrix},
\end{align}
\qed

\paragraph{Part (b)}

For Part b we assume that $\bm{R}$ is centered, so
\begin{equation}
    R_j =
    \begin{cases}
    1 - p_j & \text{w.p.~~} p_j \\
    -p_j & \text{w.p.~~} 1 - p_j
    \end{cases}.
\end{equation}
Like in the proof of (a), we want to show that 
$\E[\bm{D} \bm{D}^T]$ is $2 \times 2$ is $2 \times 2$ block diagonal.  
The only difference from the proof used in part (a) is that $\bm{R}$ is now centered. 
$\E[\tilde{Z}_j R_k] = 0$ when $j \neq k$ because $X_j$ and $R_k$ are still independent under the self-masking mechanism, and when $j = k$, 
\begin{align}
\E[\tilde{Z}_j R_j] &= \E[\tilde{Z}_j R_j \mid R_j = 1 - p_j]p_j + \E[\tilde{Z}_j R_j \mid R_j = -p_j] (1 - p_j) \\
&= p_j(1 - p_j) \left(\E[\tilde{Z}_j \mid R_j = 1 - p_j] - \E[\tilde{Z}_j \mid R_j = -p_j]\right) \\
&= 0
\end{align}
where $\E[\tilde{Z}_j \mid R_j = 1 - p_j] = 0$ because of 0 imputation and $\E[\tilde{Z}_j \mid R_j = -p_j] = 0$ because of centering. 

Different from (a), though, $R_j \notindep Y$ now because both depend on $X_j$. We thus have
\begin{align}
\E[R_j Y] &= \E[Y R_j \mid R_j = 1 - p_j] p_j + \E[Y R_j \mid R_j = -p_j] (1 - p_j) \\
&= p_j(1 - p_j) \left(\E[Y \mid R_j = 1 - p_j] - \E[Y \mid R_j = -p_j]\right).
\end{align}
Lastly, $\E[R_j, R_k] = 0$ when $j = k$ since $R_j \indep R_k$ under self-masking, and $\E[R_j^2] = p_j(1-p_j)$.
Putting this together, we have
\begin{align}
\begin{bmatrix}
\bm{\beta}^*_{MIM} \\
\bm{\gamma}^*_{MIM}
\end{bmatrix} &= 
\begin{bmatrix}
\E[\tilde{\bm{Z}}\tilde{\bm{Z}}^T]^{-1} & 0 \\
0 & \E[\bm{R} \bm{R}^T]^{-1}
\end{bmatrix}
\begin{bmatrix}
\E[\tilde{\bm{Z}} Y] \\
\E[\bm{R} Y]
\end{bmatrix} \\
&= 
\begin{bmatrix}
\beta^*_1 \\
\vdots \\
\beta^*_p \\
\E[Y \mid R_1 = 1 - p_1] - \E[Y \mid R_1 = -p_1] \\
\vdots \\
\E[Y \mid R_p = 1 - p_p] - \E[Y \mid R_p = -p_p]
\end{bmatrix}.
\end{align}
\qed

\paragraph{Part (c)}
The proof of part (c) starts by reordering the columns of $\bm{D} = [\tilde{\bm{Z}}^T, \bm{R}^T]^T$ based on the blocks $B_1, \ldots, B_d$, i.e. 
\begin{equation}
\bm{D} \coloneqq [\bm{D}_{B_1}^T, \ldots, \bm{D}_{B_d}^T]^T \coloneqq [\tilde{\bm{Z}}_{B_1}^T, \bm{R}_{B_1}^T, \ldots, \tilde{\bm{Z}}_{B_d}^T, \bm{R}_{B_d}^T]^T.
\end{equation}
Using the same logic as in the proof of part (b), we can conclude that $\E\left[ \bm{D} \bm{D}^T \right]$ is $d \times d$ block diagonal, where $d$ is the number of blocks, and further for block $B_\ell$
\begin{align}
\begin{bmatrix}
\bm{\beta}^*_{MIM_{B_\ell}} \\
\bm{\gamma}^*_{MIM_{B_\ell}}
\end{bmatrix}
= \E[\bm{D}_{B_\ell} \bm{D}_{B_\ell}^T]^{-1} \E[\bm{D}_{B_\ell} Y].
\end{align}
Recall from the proof of (b) that
\begin{equation}
\E[R_j Y] = p_j(1 - p_j) \left(\E[Y \mid R_j = 1 - p_j] - \E[Y \mid R_j = -p_j]\right)
\end{equation}
thus for block $B_\ell$
\begin{align}
&\E[\bm{D}_{B_\ell} \bm{D}_{B_\ell}^T]^{-1} \E[\bm{D}_{B_\ell} Y] \\
&= \E[\bm{D}_{B_\ell} \bm{D}_{B_\ell}^T]^{-1} 
\begin{bmatrix}
\E[\tilde{\bm{Z}}_{B_\ell} Y] \\
\E[Y \mid \bm{R}_{B_\ell} = 1 - \bm{p}_{B_{\ell}}] - \E[Y \mid \bm{R}_{B_\ell} = -\bm{p}_{B_\ell}]
\end{bmatrix}
\end{align}
where we've abused notation and used $\E[Y \mid \bm{R}_{B_\ell} = 1 - \bm{p}_{B_\ell}]$ to mean the vector of $\E[Y \mid R_j = 1 - p_j]$ for $j \in B_\ell$ and similarly for $\E[Y \mid \bm{R}_{B_\ell} = - \bm{p}_{B_\ell}]$.
\qed

\section{Additional MIM Details}\label{sec:other_cons_app}

\subsection{MIM with decision trees} 
\label{sec:mim_trees_appendix}

Supervised learning models based on decision trees are discrete models, making them different from continuous supervised learning models, like linear models and neural networks.
When decision trees split on categorical features, the split is chosen based on the levels of the categorical feature.
With numerical features, however, decision trees must choose a collection of threshold points. 
For each threshold point $T$, the tree can split on the numerical feature if it is greater than $T$ or less than $T$.
Exact details of how decision tree algorithms handle numerical features depends on the implementation; see, for example, the popular C4.5 decision tree algorithm \citep{quinlan2014c4}.

When doing imputation as a preprocessing step for tree-based models like XGBoost, the choice of imputation can significantly affect the performance of the model. 
In particular, if the imputation method is constant, i.e. always imputes the same value, then the branches will be split by these imputed values each tree with high probability. 
With an imputation method that is non-constant, e.g. GC or MF, this is no longer the case. Additionally, for non-decision-tree-based supervised models, constant and non-constant imputation can have similar behavior in some cases, e.g. using constant imputation plus noise.

Now consider using MIM in conjunction with imputation for tree-based models.
Each new indicator feature gives the tree 1 extra split to consider.
Specifically, the indicator features allow a tree to split directly on whether or not a feature is observed. 
If constant imputation is used, e.g. mean imputation, these indicator features add little additional value to the model, since all imputed values are already split together by the tree.
On the other hand, if a non-constant imputation is used, e.g. GC or MF, then the tree has much higher probability to split on the missing indicators, which can potentially really enhance the model. This explains why, throughout the experiments, MIM improves the performance of GC and MF more than the performance of Mean for experiments with XGBoost.

Since decision trees are discrete models, there are also other preprocessing methods unique to trees for dealing with missing values.
One notable example is the Missing Incorporate as Attribute (MIA) strategy \citep{twala2008good, josse2019consistency, perez2022benchmarking}, which is similar in spirit to MIM.
To illustrate MIA, let $x$ be a numerical feature to split, with missing values represented as $*$. 
For a given threshold $T$, a normal decision tree would consider the split $x \geq T$ versus $x < T$. 
MIA instead considers the following 3 splits for each split:
\begin{itemize}
    \item ($x \leq T$ or $x = *$) versus $x > T$
    \item $x \leq T$ versus ($x > T$ or $x = *$)
    \item $x = *$ versus $x \neq *$
\end{itemize}
MIA allows trees to split on missing values while doing any imputation, thereby allowing tree to utilize informative signal from the missing values if such signal is present.
Note that the third split in the above list is the same as splitting on indicator features from MIM.
The first and second splits in the list essentially compute the optimal constant imputation value for each feature, rather than always using the same value, e.g. the mean.
Thus, MIA is very similar to Mean + MIM, although using another non-constant imputer with MIM is still quite different than MIA.
A further discussion of MIA and other methods for handling missing values in decision trees can be found in \citep{josse2019consistency}.

\subsection{MIM with categorical features}
\label{sec:cat_features}

In our experiments, we consider data sets only with numeric variables, for several reasons.  First, many imputation methods cannot handle categorical variables (e.g. gcimpute), making comparisons more difficult.  Second, mean imputation is not possible with categorical variables.  The closest comparison is perhaps mode imputation, i.e. imputing with the most frequent category, but this has different properties than mean imputation and complicates the analysis. On the other hand, MIM has a very straightforward implementation for categorical variables: replace missing categorical values with a new ``missing'' category.  After one-hot encoding, this corresponds exactly to adding a new indicator column as in the numeric case.  For neural network models, this transformation could also correspond to learning a new embedding for the missing category.

\begin{table*}
    \centering
    \caption{Details of the 17 clinical features used for MIMIC experiments in Section \ref{sec:mimic}. For each feature and each task, the missing rate is given, along with whether the missingness is informative or not, based on the SMIM procedure in Algorithm \ref{alg:smim_alg}.
    Multiple features for each task have informative missingness, suggesting that MIM will likely improve model performance, which turns out to be the case in Figure \ref{fig:mimic_plot}.
    }
    \scalebox{0.95}{
    \begin{tabular}{lcccccc}
\toprule
                                   & \multicolumn{2}{c}{Mortality} & \multicolumn{2}{c}{LOS} & \multicolumn{2}{c}{Phenotyping} \\
                           Feature & Missing Rate & Is Informative & Missing Rate & Is Informative & Missing Rate & Is Informative \\
\midrule
             Capillary refill rate &        0.984 &             no &        0.973 &            yes &        0.985 &            yes \\
          Diastolic blood pressure &        0.012 &             no  &        0.069 &            yes &        0.055 &            yes \\
          Fraction inspired oxygen &        0.710 &            yes &        0.698 &            yes &        0.753 &            yes \\
    Glascow coma scale eye opening &        0.010 &             no &        0.070 &            yes &        0.055 &            yes \\
 Glascow coma scale motor response &        0.010 &             no & 0.070 &            yes &        0.056 &            yes \\
          Glascow coma scale total &        0.422 &             no &        0.438 &            yes &        0.477 &            yes \\
Glascow coma scale verbal response &        0.010 &             no &        0.070 &            yes &        0.056 &            yes \\
                           Glucose &        0.001 &             no &        0.007 &            yes &        0.011 &            yes \\
                        Heart Rate &        0.012 &             no &          0.068 &            yes &        0.054 &            yes \\
                            Height &        0.829 &            yes &        0.805 &            yes &        0.797 &            yes \\
               Mean blood pressure &        0.012 &             no &      0.069 &            yes &        0.056 &            yes \\
                 Oxygen saturation &        0.006 &             no &       0.059 &            yes &        0.048 &            yes \\
                  Respiratory rate &        0.012 &             no &        0.068 &            yes &        0.055 &            yes \\
           Systolic blood pressure &        0.012 &             no &     0.069 &            yes &        0.055 &            yes \\
                       Temperature &        0.022 &             no &     0.077 &            yes &        0.063 &            yes \\
                            Weight &        0.235 &             no &      0.223 &            yes &        0.248 &            yes \\
                                pH &        0.104 &            yes &       0.102 &            yes &        0.170 &            yes \\
\bottomrule
\end{tabular}
    }
    \label{tab:mimic_features}
\end{table*}

\begin{figure*}
    \centering
    \includegraphics[width=0.9\linewidth]{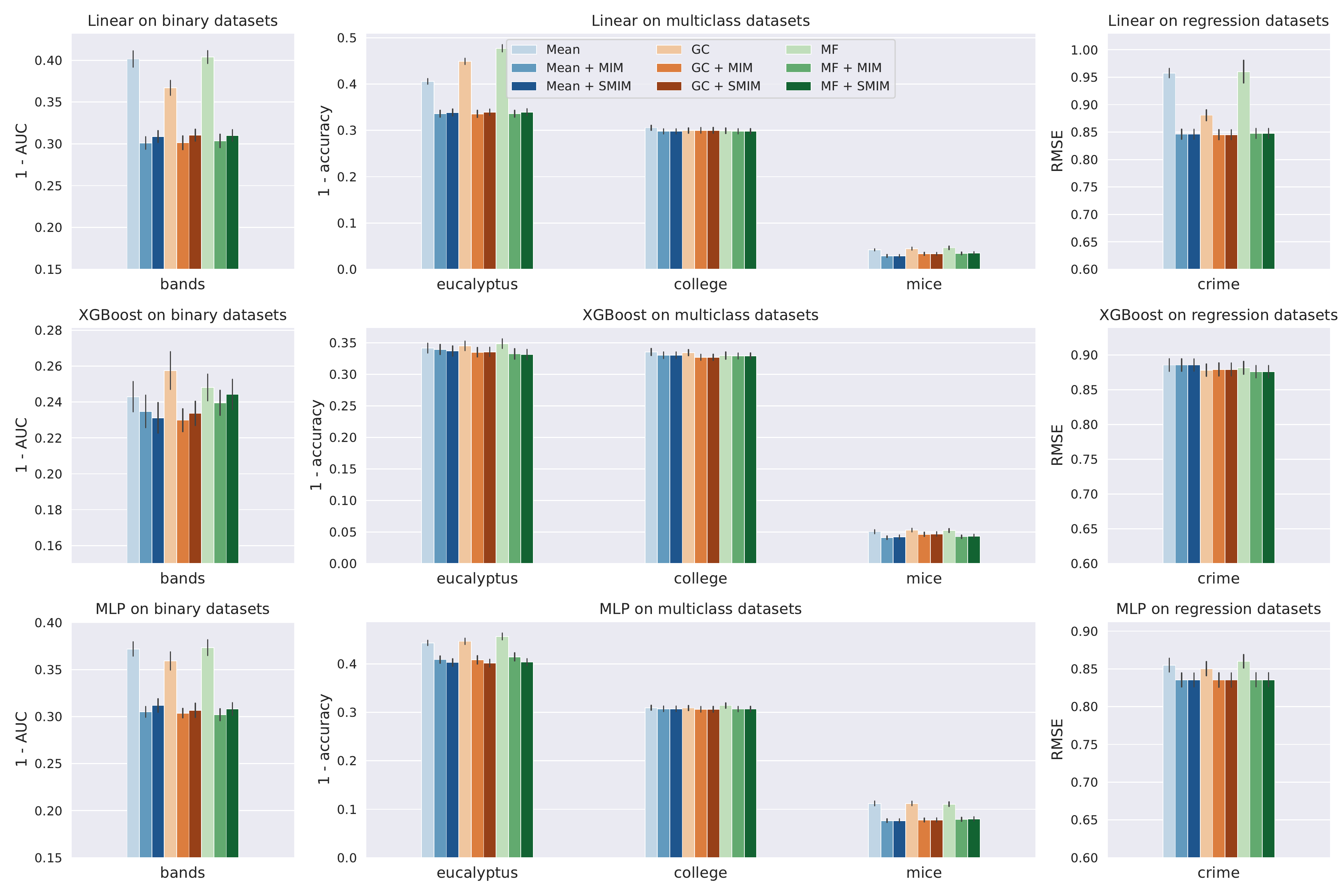}
    \caption{
    MIM and SMIM performance on OpenML data sets that already have missing values.
    Both MIM and SMIM improve linear and MLP performance on 4 out of 6 data sets, showing that missing values in real-world data often has informative missingness.
    }
    \label{fig:openml_realmiss}
\end{figure*}

\section{Additional Experiment Details}

\begin{table*}
    \centering
    \caption{OpenML data sets used.  Note that for volkert and christine, we remove several features that are either all 0 or are all 0 except a few rows, resulting in the dimensions listed. }
    \label{tab:openml_datasets}
    \vspace{0.5pc}
    \begin{tabular}{llllll}
        \toprule
        OpenML ID & Name & n & p & Task & n\_classes \\
        \midrule
        23512 & higgs & 98050 & 28 & classification & 2 \\
        1458 & arcene & 200 & 10001 & classification & 2 \\
        41150 & miniboone & 130064 & 50 & classification & 2 \\
        41145 & philippine & 5832 & 309 & classification & 2 \\
        41142 & christine & 5418 & 1599 & classification & 2 \\
        1489 & phoneme & 5404 & 5 & classification & 2 \\
        6332 & bands & 540 & 16 & classification & 2 \\
        41166 & volkert & 58310 & 147 & classification & 10 \\
        40498 & wine & 4898 & 11 & classification & 7 \\
        41163 & dilbert & 10000 & 2000 & classification & 5 \\
        188 & eucalyptus & 736 & 9 & classification & 5 \\
        488 & college & 1161 & 6 & classification & 3 \\
        537 & housing & 20640 & 8 & regression & NA \\
        42705 & yolanda & 400000 & 100 & regression & NA \\
        507 & space & 3107 & 6 & regression & NA \\
        315 & crime & 1994 & 25 & regression & NA \\
        \bottomrule
    \end{tabular}
\end{table*}

\subsection{OpenML Data Sets}
\label{sec:openml_dataset_stats}

In Table \ref{tab:openml_datasets} we show a description of the OpenML data sets used.  
The source data sets can easily by found by searching the IDs on OpenML.
We chose these data sets to represent diversity in $n$, $p$, and outcome type.
We also restrict to data sets with continuous features, as explained in Appendix \ref{sec:cat_features}.
For Figure \ref{fig:openml_main}, we only use data sets which have no missing values, since we want to be able to completely control the missing mechanism in the data.
The listed values for $p$ are the number of features used in the data sets for our experiments, which in some cases is different than the listed number of features on OpenML.
For example, there are several features in the volkert data set which all entirely 0, and so we remove these features.

\subsection{OpenML Real Missing Experiments}
\label{sec:openml_realmiss}

We show in Figure \ref{fig:mimic_plot} that MIM improves performance for EHR prediction tasks, where missing values occur naturally in the data.
We also experiment on some additional OpenML data set which already have missing values without masking.
We follow the same experimental setup as in Section \ref{sec:experiments}, except that the missing values remain the same in each trial since we are not using a synthetic missing value mask.
The results are shown in Figure \ref{fig:openml_realmiss}.
Like the results on the MIMIC tasks, MIM and SMIM improve performance on all data sets except OpenML's `college' data set for linear and MLP models.
This provides further evidence that missing values are commonly informative in real-world data sets.

\subsection{MIMIC}
\label{sec:mimic_details}

The MIMIC-III data set \cite{johnson2016mimic} is a standard data set for building models on electronic health records (EHRs), and has been used by many papers to evaluate machine learning models \cite{scherpf2019predicting, gentimis2017predicting, zhu2021machine, beaulieu2018mapping, nuthakki2019natural, ding2021artificial}. 
Due to its popularity, many tools have been developed to preprocess the raw MIMIC data into forms suitable for data science \cite{Harutyunyan2019, wang2020mimic, purushotham2018benchmarking, tang2020democratizing}.
We chose to use the mimic3-benchmark \cite{Harutyunyan2019} to help preprocess our data, creating data sets for the mortality, length of stay (LOS), and phenotype prediction tasks described in Section \ref{sec:mimic}.

For each of the above tasks, the mimic3-benchmark code gathers data from 17 clinical variables as features for the prediction. 
These 17 features are the same for each task.
The feature names are provided in Table \ref{tab:mimic_features}.
For each of the above tasks, the mimic3-benchmark code provides scripts for generating multivariate time series data, with 1 time series per feature per visit that covers a patient's data across their hospital stay.
The benchmark code also provides an additional script to generate tabular data using feature engineering on the multivariate time series data to support their logistic regression baselines.
Since we study tabular data in this paper, we use this additional preprocessing, and generate 1 feature for each clinical feature in Table \ref{tab:mimic_features} corresponding to the mean value across the observed components of the time series.
When a time series has no observed time steps, we leave the tabular feature as missing.
For each of the resulting features, we compute the missing rate and whether or not the feature has informative missingness based on SMIM, and display the results in Table \ref{tab:mimic_features}.

\end{document}